\documentclass{article}[12]

\usepackage{PRIMEarxiv}
\usepackage[utf8]{inputenc} 
\usepackage[T1]{fontenc}    
\usepackage{hyperref}       
\usepackage{url}            
\usepackage{booktabs}       
\usepackage{amsfonts}       
\usepackage{nicefrac}       
\usepackage{microtype}      
\usepackage{lipsum}
\usepackage{fancyhdr}       
\usepackage{graphicx}       
\usepackage{amsmath}
\usepackage{dsfont}

\DeclareMathOperator*{\argmax}{arg\,max}
\DeclareMathOperator*{\argmin}{arg\,min}

\usepackage{algorithmic}
\usepackage{algorithm}

\title{How optimal transport can tackle gender biases in multi-class neural-network classifiers  for job recommendations?}

\author{
Fanny Jourdan$^{1,2}$, Titon Tshiongo  Kaninku$^{1,3}$, Nicholas Asher$^{2}$, Jean-Michel Loubes$^{1}$, Laurent Risser$^{1}$\\
$\null$\\
$\null^{1}$ Institut de Math\'ematiques de Toulouse (UMR 5219), CNRS, Université de Toulouse, F-31062 Toulouse, France\\
  $\null^{2}$ Institut de Recherche en Informatique de Toulouse (UMR 5505), CNRS, Université de Toulouse, F-31062 Toulouse, France\\
  $\null^{3}$ AKKODIS group, France\\
}

\begin{document}
\maketitle

\begin{abstract}
Automatic recommendation systems based on deep neural networks have become extremely  popular during the last decade. Some of these systems can however be used for applications which are ranked as High Risk by the European Commission in the A.I. act, as for instance for online job candidate recommendation. When used in the European Union, commercial AI systems for this purpose will then be required to have to  proper statistical properties with regard to potential discrimination they could engender. This motivated our contribution, where we present a novel optimal transport strategy to mitigate undesirable algorithmic biases in multi-class neural-network classification. Our stratey is model agnostic and can be used on any multi-class classification neural-network model.
To anticipate the certification of recommendation systems using textual data, we then used it on the Bios dataset, for which the learning task consists in predicting the occupation of female and male individuals, based on their LinkedIn biography. Results show that it can reduce undesired algorithmic biases in this context to lower levels than a standard strategy.
\end{abstract}

\keywords{fairness; algorithmic bias; neural-networks; NLP; recommender systems; multi-class classification; certification}

\section{Introduction}\label{sec:intro}

The field of Artificial Intelligence (AI) has experienced remarkable growth over the past decade, particularly in Natural Language Processing (NLP). Current state-of-the-art NLP applications, such as translation or text-based recommendations, rely heavily on Deep Neural Networks (DNNs), which use transformer block layers \cite{vaswani2017attention}.
The two most widely-used transformer neural network architectures for these tasks are BERT \cite{devlin2018bert} and GPT \cite{radford2018improving}, and there are numerous variants of these models. Compared with their predecessors, such as LSTM models \cite{sutskever2014sequence}, they exhibit significantly higher performance in NLP applications. However, due to the large number of parameters and non-linearities involved, they are even less interpretable than more classic models and are typically just treated as black-box decision systems.
We developed the methodologyin this paper with the aim of controlling  undesirable algorithmic biases on recommendation systems that are built on textual information from personal profiles on social networks, such as job or housing offers. Throughout this paper, we emphasize the importance of ethical considerations in the development and deployment of these applications, as they can significantly impact users' lives.

For a long time, many believed that machine learning algorithms couldn't be discriminatory since they lack human emotions. This view is however outdated now, as different studies have shown that an algorithm can learn and even amplify biases from a biased dataset \cite{BesseEtAl_AmStat2021}. In this paper, we use the term \emph{algorithmic biases} to refer to  automatic decisions made by machine learning algorithm that are not neutral, fair, or equitable for a particular subgroup of people (or statistical observations in general). This group is distinguished by a \emph{sensitive variable}, such as gender, age, or ethnic origin. The field of study and prevention of these specific algorithmic biases is called \emph{Fair learning}.
Ensuring fairness is essential to ensure an ethical application of algorithms in society.  Ethical concerns have become increasingly important in recent years, and the deployment of a discriminatory algorithm is no longer acceptable. Many regulators are already addressing ethical issues related to AI. In the area of privacy, the General Data Protection Regulation (GDPR), which was adopted by the European Parliament in 2016, allows for instance to the French Commission on Informatics and Liberty (NCIL) and other independent administrative authorities in France to impose severe penalties on companies that do not manage customer data transparently \cite{de2018titre}. The GDPR is an example of how public authorities are progressively developing legal frameworks and taking actions to mitigate threats. More recently, the so-called \emph{AI act} \footnote{\url{https://eur-lex.europa.eu/legal-content/EN/TXT/?uri=CELEX\%3A52021PC0206}} of the European Commission defined a list of \emph{High Risk} applications of A.I., most of them being related to a strong impact on the Humans life. For instance, job candidates recommendation systems are therefore ranked as  \emph{High Risk}. Importantly, when sold in or from the European-Union, such A.I.  systems will be need to have appropriate statistical properties with respect to any potential discrimination they may cause (see articles 9.7, 10.2, 10.3, and 71.3).

Motivated by the future certification of A.I. systems based on black-box neural-networks against discrimination, our article expands on the work of \cite{risser2022tackling} to address algorithmic biases observed in NLP-based multi-class classification. The main methodological novelties of this paper are the extension of \cite{risser2022tackling} to multi-class classification, and to showcase how to apply it to NLP data in an application ranked as \emph{High Risk} by the \emph{AI act}. The bias mitigation model proposed in this paper involves incorporating a regularization term, in addition to a standard loss, when optimizing the parameters of a neural network. This regularization term mitigates algorithmic bias by enforcing the similarity of prediction or error distributions for two groups distinguished by a predefined binary sensitive variable (e.g. males and females), measured using the 2-Wasserstein distance between the distributions. Note that  \cite{risser2022tackling} is the first paper that demonstrated how to calculate pseudo-gradients of this distance in a mini-batch context, enabling the use of this method to train deep neural-networks with reduced algorithmic bias.

To extend \cite{risser2022tackling} to multi-class classification with deep neural-networks, we need to address a key problem: estimating the Wasserstein-2 distance between multidimensional distributions (where the dimension equals the number of output classes) requires numerous neural-network predictions, leading to slow training.  In order to avoid this problem, we will redefine the regularization term to apply to predicted classes of interest, making the bias mitigation problem numerically feasible.
Our secondary main contribution from an end-user perspective is demonstrating how to mitigate algorithmic bias in a text classification problem using modern transformer-based neural network architectures like RoBERTa small \cite{liu2019roberta}. It is important to note that our regularization strategy is model-agnostic and could be applied to other text classification models, like those based on LSTM architectures. We evaluate our method using the \emph{Bios} dataset \cite{de2019bias}, which includes over 400,000 LinkedIn biographies, each with an occupation and gender label. This dataset is commonly used to train automatic recommendation models for employers to select suitable candidates for a job and quantify algorithmic biases in the trained models. The \textit{Bios} dataset is a key resource for the scientific community studying algorithmic bias in NLP.

\section{Definitions and related work}\label{sec:relatedWork}

\subparagraph{Measuring algorithmic biases in machine learning}
Different popular metrics exist to measure algorithmic biases in the machine learning literature.
In this paper, we will use the True Positive Rate gap (TPRg) \cite{de2019bias}, which is one of the classic fairness metrics for NLP. Other metrics such as the Statistical Parity \cite{dwork2012fairness} or Equalized Odds \cite{hardt2016equality} are also very popular. Over 20 different fairness metrics are compared in \cite{verma2018fairness}. Very importantly for us, each metric shows specific algorithmic bias properties and not all of them are compatible with each other \cite{kleinberg2016inherent, chouldechova2017fair, pleiss2017fairness}.
For instance, the True Positive Rate gap quantifies the difference between the portion of positive predictions ($\hat{Y}=1$ using common M.L. notations) in two groups, by only considering the observations which should be classified as positive ($Y=1$  using common M.L. notations). Another popular metric such as the disparate impact will also quantify this difference, but for all observations. This makes their practical interpretation different.

\subparagraph{Impact of AI biases in society}
The use of artificial intelligence (AI) in decision-making systems has become increasingly widespread in recent years, and with it, concerns over the potential for discriminant bias to affect the outcomes of these decisions.
We review below different key studies that have explored the impacts of such bias in AI on society.
One such study  \cite{skeem2015risk} focused on the criminal justice system, and found that AI algorithms can produce biased outcomes, particularly when trained on non-representative data sets. This can result in higher incarceration rates for certain groups, such as racial minorities, and perpetuate systemic racism in the criminal justice system.
Another study by \cite{cirillo2020sex} explored how gender differences and biases can affect the development and use of artificial intelligence in the field of bio-medicine and healthcare. The paper discussed the potential consequences of these differences and biases, including unequal access to healthcare and inaccurate medical diagnoses.
Another important impact of algorithmic biases in society is the case of online advertisements. Ad targeting based on demographic factors such as race, gender, and age rather than interests or behaviours can perpetuate negative stereotypes and result in discrimination by limiting access to job or housing announcements for certain groups. For example, Facebook's ad delivery algorithms can lead to biased outcomes by optimizing for maximum engagement, which can result in the amplification of certain groups or messages over others. This can lead to discrimination against certain groups, as advertisers may target their ads to specific demographics or exclude certain groups from seeing their housing and employment advertising, as highlighted by studies such as \cite{ali2019discrimination} and \cite{sapiezynski2019algorithms}.

\subparagraph{Bias mitigation in NLP}

Bias in NLP systems has received a significant attention in recent years, with researchers and practitioners exploring various methods for mitigating bias in NLP models. In this subsection, we review some of the existing work on bias mitigation in NLP.
The first approach to mitigate bias is to apply \emph{pre-processing} techniques to the data used to train the model. Some researchers have proposed methods for removing or neutralizing sensitive attributes from the training data, such as gender or race, in order to reduce the likelihood that the model will learn to make decisions based on these attributes. We can reduce the bias directly in the text of the training dataset. For example, in the case of gender bias like the study in this paper, the most classic technique is to remove explicit gender indicators \cite{de2019bias}. This technique is the one we will use to compare our proposed strategy to another one commonly used in industry. This technique is indeed simple to implement and makes it possible to reduce the bias, but in a partial and not very localized manner.
Other classical techniques can be used, like identifying biased data in word embeddings, which represent words in a vector space. \cite{garg2018word} demonstrated that these embeddings reflect societal biases. There are also methods to show how these embeddings can be unbiased by aligning them with a set of neutral reference vectors \cite{zhao2018gender, caliskan2017semantics}.  These de-biasing methods have however strong limitations, as explained in \cite{gonen2019lipstick}, where the authors show that although the de-biasing embedding methods can reduce the visibility of gender bias, they do not completely eliminate it.

A second approach is to use \emph{post-processing} de-biasing methods. These methods are model-agnostic and therefore not specific to NLP since they modify the results of previously trained classifiers in order to achieve fairer results.  \cite{hardt2016equality, sikdar2022getfair} investigate this for binary classification, and \cite{denis2021fairness} propose a method for multiclass classification.

The last approach to mitigate biases in AI is to use fairness-aware algorithms, which are specifically designed to take into account the potential for bias and to learn from the data in a way that reduces the risk of making biased decisions.  These are the \emph{in-processing} methods, which generally do not depend on the type of data input, either. The method we propose in this paper is one of them. To achieve this,  we can use adversarial Learning by adjusting the discriminator. Adversarial learning involves training a model to make predictions while also training a second model to identify and correct any biases in the first model's predictions. By incorporating this technique into the training process, \cite{zhang2018mitigating, madras2018learning} demonstrate that it is possible to reduce the amount of bias present in machine learning models. Another technique is to constrain the predictions with a regularization technique like \cite{kamishima2012fairness}, but this technique was only used on a logistic regression classifier. On the other hand, \cite{manisha2018fnnc} mitigate fairness specifically in neural networks.
Finally, \cite{zafar2017fairness, zafar2017fairness2} use fairness metrics constraints and solve the training problem subject to those constraints.
All these \textit{in-processing} methods apply in the case of binary classification. There is indeed an \textit{in-processing} paper that proposes a method for multiclass classification for a computer vision task \cite{zhao2017men}, but this paper focuses on the regularization of the mean bias amplification and therefore does not deal with the classic fairness metrics.

We want to emphasize that the \emph{pre-processing} and \emph{post-processing} methods do not reduce the bias to the same degree across the whole machine-learning training procedure than \emph{in-processing} methods. Our paper hencefocuses on an \emph{in-processing} method.
In this context, our methodology tackles an issue which was still not addressed in the fair learning literature, as far as the authors know: we tackle algorithmic biases on multi-class neural-network classifiers and not on binary classifiers or on non neural-network classifiers. We believe that the potential of such a strategy is high for the future certification of commercial A.I. systems.

\section{Methodology}

The bias mitigation technique proposed in this paper extends the regularization strategy of \cite{risser2022tackling} to multi-class classification. In this section, we first introduce our notation, then describe the regularization strategy of \cite{risser2022tackling} for binary classifiers, and then extend it to multi-class classifiers. This extension is the methodological contribution of our manuscript.

\subsection{General notations}\label{ssed:GeneralNotations}

\subparagraph{Input and output observations}
Let $(x_i, y_i)_{i = 1,\dots, n}$ be the training observations, where $x_i \in \mathbb{R}^p$ and $y_i \in \{0,1\}^K$ are the input and output observations, respectively. The value $p$ represents the inputs dimension or equivalently the number of input variables. It can for instance represent a number of pixels if $x_i$ is an image or a number of words in a text if $x_i$ is a word embedding. The value $K$ represents the output dimensions. In a binary classification context, \emph{i.e.} if $K=1$, the fact that $y_i=0$ or $y_i=1$ specifies the class of the observation $i$. In a multi-class classification context, \emph{i.e.} if $K>1$, a common strategy consists in using one-hot vectors to encode the class $c$ of observation $i$: All values $y_i^k$, $k \in \{1,\ldots,K\}$ are equal to $0$, except the value $y_i^{c}$, which is equal to $1$. We will use this convention all along this manuscript.

\subparagraph{Prediction model}
A classifier $f_{\theta}$ with parameters $\theta$ is trained so that the predictions $\hat{y_i} \in \{0,1\}^K$ it indirectly makes based on the outputs $f_{\theta}(x_{i}) \in [0,1]^K$, are \emph{in average}
as close as possible to the true output observations $y_i$ in the training set.
The link between  the model outputs $f_{\theta}(x_{i})$ and the prediction $\hat{y_i}$ depends on the classification context:
In binary classification,  $f_{\theta}(x_{i})$ is the predicted probability that $\hat{y_i}=1$, so it is common to use $\hat{y_i}=1_{f_{\theta}(x_{i})>0.5}$. Now, by using one-hot-encoded output vectors in multi-class classification, an output $f_{\theta}(x_{i}) = \left( f_{\theta}^1(x_{i}) , f_{\theta}^2(x_{i}), \ldots ,f_{\theta}^K(x_{i}) \right)$ represents the predicted probabilities that the observation $i$ is in the different classes $k \in \{1,2,\ldots,K\}$.
As a consequence, $\sum_k f_{\theta}^k(x_{i}) = 1$. More interestingly for us,
the predicted class is the one having the highest probability, so $\hat{y_i}$ is a vector of size $K$ with null values everywhere, except at the index $\argmax_k f_{\theta}^k(x_{i})$, where its value is $1$.

\subparagraph{Loss and empirical risk}
In order to train the classification model, an empirical risk $\mathcal{R}$ is minimized with respect to the model parameters $\theta$
\begin{equation}\label{eq:empiricalRisk}
\mathcal{R}(\theta):= \mathbb{E}[\ell (\hat{Y}:=f_{\theta} (X), Y)] \,,
\end{equation}
or empirically $R(\theta) = \frac{1}{n} \sum_{i=1}^n \ell (\hat{y_i}:=f_{\theta} (x_i), y_i)$, where the loss function $\ell$ represents the price paid for inaccuracy of predictions. This optimization problem is almost systematically solved by using variants of the stochastic (or mini-batch) gradient descent
\cite{Bottou_SIAMrev2018} in the machine learning literature.

\subparagraph{Sensitive variable}
An important variable in the field of \emph{fair learning} is the so-called \emph{sensitive variable}, which we will denote $S$. This variable is often binary and distinguishes two groups of observations $S_i \in \{0,1\}$. For instance, $S_i=0$ or $S_i=1$ can indicate that the person represented in observation $i$ is either a male or a female.
A widely used strategy to quantify that a prediction model is fair with respect to the variable $S$ is to compare the predictions it makes on observations in the groups $S=0$ and $S=1$, using a pertinent \emph{fairness metric} (see references of Section \ref{sec:relatedWork}). From a mathematical point of view, this means that the difference between the distributions $(X,Y,\hat{Y})_{S=0}$ and $(X,Y,\hat{Y})_{S=1}$, quantified by the fairness metric,  should be below a given threshold. Consider for instance a binary prediction case where  $\hat{Y_i}=1$ means that the individual $i$ has access to a bank loan, $\hat{Y_i}=0$ means that the bank loan is refused, and that $S_i$ equals $0$ or $1$ refers to the fact that the individual $i$ is  a male or a female. In this case, one can use the difference between the empirical probabilities of obtaining the bank loan for males and females, as a fairness metric, \textit{i.e.} $P(\hat{Y}=1 | S=1) - P(\hat{Y}=1 | S=0)$. More advanced metrics  may also take into account the input observation $X$, the true outputs $Y$, or the prediction model outputs $f_{\theta}(X)$ instead of their binarized version $\hat{Y}$.

\subsection{W2reg approach for binary classification}

\subsubsection{Regularisation strategy}

We now give an overview of the \emph{W2reg} approach, described in \cite{risser2022tackling}, to temper algorithmic biases of binary neural-network classifiers. The goal of  \emph{W2reg} is to ensure that the treated binary classifier $f_{\theta}$ generates predictions $\hat{Y}$ for which the  distributions in groups $S=0$ and $S=1$ do not deviate too much from pure equality.
To achieve this, the similarity metric used in \cite{risser2022tackling} is the Wasserstein-2 distance between the distribution of the predictions in the two groups:
\begin{equation}\label{eq:w2regstar}
\mathcal{W}_2^2 (\mu_{\theta,0},\mu_{\theta,1}) = \int_0^1   \left( {\mathcal{H}_{\theta,0}}^{-1} (\tau) - {\mathcal{H}_{\theta,1}}^{-1} (\tau) \right)^2 d \tau \,.
\end{equation}
where $\mu_{\theta,s}$ is the probability distribution of the predictions made by $f_{\theta}$ in group $S=s$, and ${\mathcal{H}_{\theta,s}}^{-1}$ is the inverse of the corresponding  cumulative distribution function.
Note that $\mu_{\theta,s}$ is mathematically equivalent to the histogram of the model outputs $f_{\theta}(X)$ for an infinity of observations in the group $S=s$, after normalization so that the histogram integral is $1$.
Remark, that this metric is also based on the model outputs $f_{\theta}(X) \in [0,1]$ and not the discrete predictions $\hat{Y} \in \{0,1\}$ (see Section \ref{ssed:GeneralNotations}-\textit{Prediction model} for the formal relation), so the probability distributions $\mu_{\theta,s}$ are continuous.
Ensuring that this metric remains low makes it possible to control the level of fairness of the neural-network model $f_{\theta}$ with respect to $S$. As specifically modelled by Eq.~\eqref{eq:w2regstar}, this is made by penalizing the average squared difference between the quantiles of the predictions in the two groups.
In order to train neural-network which simultaneously make accurate and fair decisions, the strategy of  \cite{risser2022tackling} then consists in optimizing the parameters $\theta$ of the model $f_{\theta}$ such that:
\begin{equation}\label{eq:minimizedEnergy}
\widehat{\theta} = \argmin_{\theta \in \Theta} \left\{ \mathcal{R}(\theta) + \lambda \mathcal{W}_2^2 (\mu_{\theta,0},\mu_{\theta,1}) \right\} \,,
\end{equation}
where $\Theta$ is the space of the neural-network parameters (\textit{e.g.} the values of the weights, the bias terms and the convolution filters in a CNN). As usual when training a neural-network, the parameters $\theta$ are optimized using a gradient-descent approach, where the gradient is approximated at each gradient-descent step by using a mini-batch of observations.

\subsubsection{Gradient estimation}

We compute the gradient of Eq.~\eqref{eq:minimizedEnergy} using the standard back-propagation strategy \cite{LeCunNC1989}. For the empirical risk part of Eq.~\eqref{eq:minimizedEnergy}, this requires to compute the derivatives of the  losses $\ell (f_{\theta}(x_i), y_i)$ with respect to the neural-network outputs $f_{\theta}(x_i)$, which is routinely made by solutions like PyTorch, TensorFlow or Keras,  for all  mainstream losses. For the Wasserstein-2 part of Eq.~\eqref{eq:minimizedEnergy}, the authors of \cite{risser2022tackling} proposed to use a mathematical strategy to compute pseudo-derivatives of $\mathcal{W}_2^2 (\mu_{\theta,0},\mu_{\theta,1})$ with respect to the neural-network outputs $f_{\theta}(x_i)$.
Specifically, to compute the pseudo-derivative of a discrete and empirical approximation of $\mathcal{W}_2^2 (\mu_{\theta,0},\mu_{\theta,1})$ with respect to a mini-batch output $f_{\theta}(x_i)$, the following equation was used:
\begin{equation}\label{eq:discreteDerivative}
\displaystyle{
\Delta_{\tau} \left[
 \mathds{1}_{s_i=0}
 \frac{     f_{\theta}(x_i) - cor_1(f_{\theta}(x_i))   }{ n_0 \left(H_{0}^{j_i+1} - H_{0}^{j_i} \right)}
 -
 \mathds{1}_{s_i=1}
 \frac{     cor_0(f_{\theta}(x_i)) - f_{\theta}(x_i) }{ n_1 \left(H_{1}^{j_i+1} - H_{1}^{j_i} \right)}
\right]
} \,,
\end{equation}
where
$n_s$ is the number of observations in class $S=s$, the $H_s^{j}$ are discrete versions of the cumulative distribution functions $\mathcal{H}_{\theta,s}$ defined on a discrete grid of possible output values:
\begin{equation}\label{eq:PbDiscretization}
\eta^j = \min_i(f_{\theta}(x_i)) + j \Delta_{\eta} \,,\, j=1, \ldots, J_{\eta} \,,
\end{equation}
where $\Delta_{\eta} = J_{\eta}^{-1} (\max_i(f_{\theta}(x_i)) - \min_i(f_{\theta}(x_i)) )$, and $J_{\eta}$ is the number of discretization steps. We denote $H_s^j = H_s (\eta^j)$ and $j_i$ is defined such that $\eta^{j_i} \leq f_{\theta}(x_i) < \eta^{j_i+1}$. Finally  $cor_s(f_{\theta}(x_i)) = H_{s}^{-1}(H_{|1-s|}(f_{\theta}(x_i)))$.

\subsubsection{Distinction between mini-batch observations and the  observations for $H_0$ and  $H_1$}

As shown in Eq.~\eqref{eq:discreteDerivative}, computing the pseudo-derivatives of Wasserstein-2 distance $W_2^2 (\mu^n_{\theta,0},\mu^n_{\theta,1})$ with respect to model predictions $f_{\theta}(x_i)$ requires computing the discrete cumulative distribution functions $H_{s}$, with $s \in \{0,1\}$. Computing the  $H_{s}$ would ideally require computing $f_{\theta}(x_i)$ for all $n$
observations $x_i$ of the training set, which would be computationally bottleneck. To solve this issue, \cite{risser2022tackling} proposed to approximate the $H_{s}$ at each mini-batch iteration, where Eq.~\eqref{eq:discreteDerivative} is computed, using a subset of all training observations. This observation subset is composed of $m$ randomly drawn observations in group $S=0$, $m$ other randomly drawn observations in group $S=1$, and the mini-batch observations. This guaranties that there are at least $m$ observations to compute either $H_0$ or $H_1$, and that the impact of each mini-batch observation is represented in $H_0$ and $H_1$. Note that these additional $2m$ predictions do not require to backpropagate any gradient information, so their computational burden is limited in terms of memory resources. Although it is also reasonable in terms of computational resources, the amount of $2m$ additional predictions should remain relatively small to avoid slowing down significantly the gradient descent. In previous experiences on images, $m=16$
or $m=32$ often appeared as reasonable, as it allowed to mitigate undesirable algorithmic biases and slowed down the whole training procedure by a factor of less than $2$.  We finally want to emphasize that preserving the amount of such additional predictions reasonable at each gradient descent step will be at the heart of our methodological contribution when extending \emph{W2reg} to multi-class classification.

\subsection{Extended W2reg for multi-class classification}

As discussed in Section \ref{sec:intro}, our work is motivated by the need for bias mitigation strategies in NLP applications where the neural-network predicts that an input text belongs to a class among more $K$ output classes, where $K>2$. We then show in this section how to take advantage of the properties of \cite{risser2022tackling} to address this practical problem. We recall that the regularization strategy of \cite{risser2022tackling} is model agnostic, so the fact that we treat NLP data will only be discussed in the results section. In terms of methodology, the main issue to tackle is that the model outputs   $f_{\theta}(x_i)$ are in dimension $K>2$ and not one dimensional, which would require to compare multi-variate point clouds following the optimal transport principles which were modelled by Eq.~\eqref{eq:w2regstar} for 1D outputs. As we will see below, this  issue opens algorithmic problems to keep the computational burden reasonable and to preserve the representativity of the pertinent information. Solving them requires extending \cite{risser2022tackling} with strong algorithmic constraints.

\subsubsection{Reformulating the bias mitigation procedure for multi-class classification}

The strategy proposed by \cite{risser2022tackling} to mitigate undesired biases is to train optimal decision rules $f_{\theta}$ by optimizing Eq.~\eqref{eq:minimizedEnergy}, where the Wassertein-2 distance between the prediction distributions $\mu_{\theta,0}$ and $\mu_{\theta,1}$ (\textit{i.e.} the distribution of the predictions $f_{\theta}$ for observations in groups $S=0$ and group $S=1$) is given by Eq.~\eqref{eq:w2regstar}.
As described in Section \ref{ssed:GeneralNotations}, the predictions $f_{\theta}(x_i)$ are now a vector of dimension $K>2$ in a multi-class classification context (specifically $f_{\theta}(x) \in [0,1]^K$). Their distributions $\mu_{\theta,0}$ and $\mu_{\theta,1}$ are then multivariate. In this context, Eq.~\eqref{eq:w2regstar} does no hold, and another optimal transport metric such as the multivariate Wasserstein-2 Distance or the Sinkhorn Divergence should be used \cite{ChizatEtAl20}. Note that different implementations of these metrics exist and are compatible with our problem, as \textit{e.g.} those of  \cite{flamary2021pot,feydy2019interpolating}.
This however opens a critical issue related to the number of observations needed to reasonably penalize the differences between two multivariate point clouds, representing the observations in groups $S=0$ and $S=1$. If the dimension $K$ of the compared data gets large, the number of observations required to reasonably compare the point clouds at each gradient descent step  explodes. This problem is very similar to the well known \emph{curse of dimensionality} phenomenon in machine learning, where the  amount of data needed to generalize accurately the predictions grows exponentially as the number of dimensions grows.

This issue therefore leads us to think about which problem we truly need to solve when tackling undesired algorithmic bias in multi-class classification. From our application perspective, a discrimination appears when there the prediction model $f_{\theta}$ is significantly more accurate to predict a specific output in one of the two groups represented by $S \in \{0,1\}$. For instance, suppose that someone looks for \textit{Software Engineer} jobs and that an automatic prediction model $f_{\theta}$ is used to recommend job candidates to an employer. For a given job candidate $x_i$, the prediction model  will return a set of $K$ probabilities, each of them indicating whether $x_i$ is recommended for the job class $k$. Now $k$ will denote the class of jobs $x_i$ is looking for, \textit{i.e.} \textit{Software Engineer}.
The prediction model will be considered as unfair if male  profiles are on average clearly more often recommended by $f_{\theta}$ than female profiles, when an unbiased oracle would lead to equal opportunities, \textit{i.e.}
\begin{equation}\label{eq:TPRg}
|P(\hat{Y}^k=1 | Y^k=1 , S=1) - P(\hat{Y}^k=1 | Y^k=1 , S=0) | > \tau \\,
\end{equation}
where the left-hand term is denoted the \emph{True Positive Rate  gap} (TPRg), and $\tau$ is a threshold above which the TPRg is considered as discriminant. As shown  in Section \ref{sec:results}, such situations can occur in automatic job profile recommendation systems using modern neural-networks.
Now that the problem to tackle is clarified, we can reformulate the regularized multi-class model training procedure as follows:
\begin{itemize}
\item
We first train and test a non-regularized multi-class classifier $f_{\theta^{bl}}$. We will denote it \emph{baseline classifier}.
\item
We define a threshold $\tau$ under which all occupation to predict $k\in \{1,\ldots,K\}$ should have a TPRg (see Eq.~\eqref{eq:TPRg}). We denote $\{c_1,\ldots,c_C\}$ the classes for which this condition is broken, where each of these classes takes its values in $\{1,\ldots,K\}$.
\item
We then retrain the multi-class classifier $f_{\theta}$ with regularization constrains on the classes $\{c_1,\ldots,c_C\}$ only. The regularization strategy will be developed below in Section~\ref{ssec:regStrategy}.
\end{itemize}

By using this procedure, the number of observations required at each mini-batch step will be first limited to observations in the groups $\{c_1,\ldots,c_C\}$ only, which is a first step towards an algorithmically reasonable regularized training procedure. We also believe that this also avoids to over-constrain the training procedure, which often penalizes its convergence.

\subsubsection{Regularization strategy}\label{ssec:regStrategy}

We now push further the algorithmic simplification of the regularization procedure by focusing on the properties of the mini-batch observations. In this subsection, we  suppose that $x_i$ is an input mini-batch observation, and recall  that we want to penalize large TPRg for specific classes $\{c_1,\ldots,c_C\}$ only. In this mini-batch step, the observations $x_i$ related to true output predictions $y_i^k=1$ for which  $k \notin \{c_1,\ldots,c_C\}$ are not concerned by the regularization, when computing the multivariate cumulative distribution function $H_0$ or $H_1$.
At each mini-batch step, it therefore appears as appealing to only consider the dimensions out of $\{c_1,\ldots,c_C\}$, for which at least one true output observation respects $y_i^k=1$, with $k \in \{c_1,\ldots,c_C\}$.
This would indeed allow to further reduce the amount of additional predictions made in the mini-batch. The dimension of $H_0$ or $H_1$ would however vary at each mini-batch step, making potentially the distance estimation unstable if fully considering $C$-dimensional distributions.

To  take into account the fact that not all output dimensions $\{c_1,\ldots,c_C\}$ should be considered at each gradient descent step, we then make a simplification hypothesis: we neglect the relations between the different dimensions when comparing the output predictions in groups $S=0$ and $S=1$. This hypothesis is the same as the one made when using Naive Bayes classifiers \cite{IdiotBayes2001,rish2001empirical}. We believe that this hypothesis is particularly suited for one-hot-encoded outputs, as they are constructed to ideally have a single value close to 1 and all other values close to 0. We then split the multi-variate regularization strategy into a multiple one-dimensional strategy and optimize:
\begin{equation}\label{eq:minimizedEnergy_extension}
\widehat{\theta} = \argmin_{\theta \in \Theta} \left\{ \mathcal{R}(\theta) +
\sum_{l=1}^{C}
\lambda_{c_l} \mathcal{W}_2^2 (\mu_{\theta,0}^{c_l},\mu_{\theta,1}^{c_l}) \right\} \,,
\end{equation}
where $\mathcal{W}_2^2$ is the metric of Eq.~\eqref{eq:w2regstar}, $\lambda_{c_l}$ is the weight given to regularize the TPR gaps in class $c_l$, and the $\mu_{\theta,s}^{c_l}$ are the distributions of the output predictions on dimension $c_l$, \textit{i.e.} the distribution of the $f_{\theta}^{c_l}(x)$, when the true prediction is $c_l$,  \textit{i.e.} when $y^{c_l}=1$.
For a mini-batch observation $x_i$ related to an output prediction in a regularized class $k \in \{c_1,\ldots,c_C\}$,
the impact of a mini-batch output $f_{\theta}^k(x_i)$ on the empirical approximation of $\mathcal{W}_2^2 (\mu_{\theta,0}^{k},\mu_{\theta,1}^{k})$ can then be estimated by following the same principles as in \cite{risser2022tackling}. We can then extend Eq.~\eqref{eq:discreteDerivative} with:
\begin{equation}\label{eq:discreteDerivative_extension}
\displaystyle{
\Delta_{\tau} \left[
 \mathds{1}_{s_i=0}
 \frac{     f_{\theta}^k(x_i) - cor_1(f_{\theta}^k(x_i))   }{ n_{k,0} \left(H_{k,0}^{j_i+1} - H_{k,0}^{j_i} \right)}
 -
 \mathds{1}_{s_i=1}
 \frac{     cor_0(f_{\theta}^k(x_i)) - f_{\theta}^k(x_i) }{ n_{k,1} \left(H_{k,1}^{j_i+1} - H_{k,1}^{j_i} \right)}
\right]
} \,,
\end{equation}
where the $H_{k,s}$ are discrete and empirical versions of the cumulative distribution functions of the prediction outputs on dimension $k$, \textit{i.e.} the  $f_{\theta}^k(x)$, when class $k$ should be predicted and the observations are in the group $s$.
Note too that Eq.~\eqref{eq:discreteDerivative} contains $n_s$, which is the number of observations in class $S=s$. In order to manage unbalanced output classes in the multi-class classification context, we also use a normalizing term $n_{k,s}$ in Eq.~\eqref{eq:discreteDerivative_extension}. It quantifies the number of training observations in group $s \in \{0,1\}$ and class $k \in \{1,\ldots,K\}$. Other notations are the same as in Eq.~\eqref{eq:discreteDerivative}.

In a mini-batch step, suppose finally that only need to take into account the classes
$\{\widehat{c_1}, \ldots , \widehat{c_{D}}\}$ among the $\{c_1,\ldots,c_C\}$. These selected classes are those for which at least a $y_i^{c_j}=1$, with $j \in \{1,\ldots,C\}$ and, $i$ is an observation of the mini-batch $B \subset \{1,\ldots,n\}$. We will then have to only sample  two times $m$ predictions, for each of the selected $D$ classes, to compute the $H_{\widehat{c_d},0}$ and $H_{\widehat{c_d},1}$ required in Eq.~\eqref{eq:discreteDerivative_extension}. This makes the computational burden to regularize the  neural-network training procedure reasonable, as the number of additional predictions to make only increases linearly with the number of treated classes at each mini-batch iteration. Note that no additional prediction will also be needed when a mini-batch contains no observation related to a regularized output class. This will naturally be often the case, when the number of classes $K$ gets large and/or the mini-batch size $\#B$ is small.

\subsubsection{Proposed training procedure}\label{ssec:regStrategy}

The proposed strategy to train multi-class classifiers with mitigated algorithmic biases on specific classes prediction was motivated by the future need of certifying that automatic decision models are not discriminatory. In order to make absolutely clear our strategy, we detail it in algorithm \ref{alg:BTP_NN_W2_mutli_class}.

\begin{algorithm}
\caption{Procedure to train bias mitigated multi-class neural-network classifiers}
\label{alg:BTP_NN_W2_mutli_class}
\begin{algorithmic}[1]
\REQUIRE Training observations $(x_i,s_i,y_i)_{i=1,\ldots,n}$, where $x_i\in \mathbb{R}^p$, $s_i \in \{0,1\}$ and $y_i \in \{0,1\}^K$, plus a multi-class neural-network model $f_{\theta}$.
\STATE [\textit{Detection of the output classes with discriminatory predictions}]
\STATE Train the baseline parameters $\theta^{bl}$ of $f$ on $(x_i,s_i,y_i)_{i=1,\ldots,n}$ with no specific regularization.
\STATE Find the output classes $\{c_1,\ldots,c_C\}$ on which the model $f_{\theta^{bl}}$ has unacceptable True Positive Rate gaps (TPRg) using Eq.~\eqref{eq:TPRg}.
\STATE [\textit{Multi-class W2reg training}]
\STATE Re-initialize the training parameters $\theta$.
\FOR{$e$ in Epochs}
\FOR{$b$ in Batches}
\STATE Draw the batch observations $(x_i,s_i,y_i)_{i \in B}$, where $B$ is a subset of $\{1,\ldots,n\}$.
\STATE Compute the mini-batch predictions $f_{\theta}(x_i)$, $i\in B$.
\STATE Detect the output classes $\{\widehat{c_1}, \ldots , \widehat{c_{D}}\}$ among the $\{c_1,\ldots,c_C\}$ for which at least a $y_i^{c_j}=1$, with $j \in \{1,\ldots,C\}$ and $i\in B$.
\STATE For each $j \in \{1,\ldots,D\}$, pre-compute $H_{\widehat{c_{j}},0}$ and $H_{\widehat{c_{j}},1}$ using $m$ output predictions $f_{\theta}(x_i)$, where $i\notin B$ and $y_i^{\widehat{c_{j}}}=1$.
\STATE Compute the empirical risk and its derivatives with respect to $f_{\theta}(x_i)$, $i\in B$.
\STATE Compute the pseudo-derivatives of the discretized $W_2^2 (\mu^{\widehat{c_j}}_{\theta,0},\mu^{\widehat{c_j}}_{\theta,1})$ with respect to the pertinent mini-batch outputs $f_{\theta}(x_i)^{\widehat{c_j}}$ using Eq.~\eqref{eq:discreteDerivative_extension}.
\STATE Backpropagate the risk derivatives and the pseudo-derivatives of the $W_2^2$ terms.
\STATE Update the parameters $\theta$.
\ENDFOR
\ENDFOR
\RETURN Trained neural network $f_{\theta}$ with mitigated biases.
\end{algorithmic}
\end{algorithm}

\section{Experimental protocol to assess W2reg on multi-class classification with NLP data}

\subsection{Data}

\begin{figure}
    \centering
\includegraphics[width=0.9\linewidth]{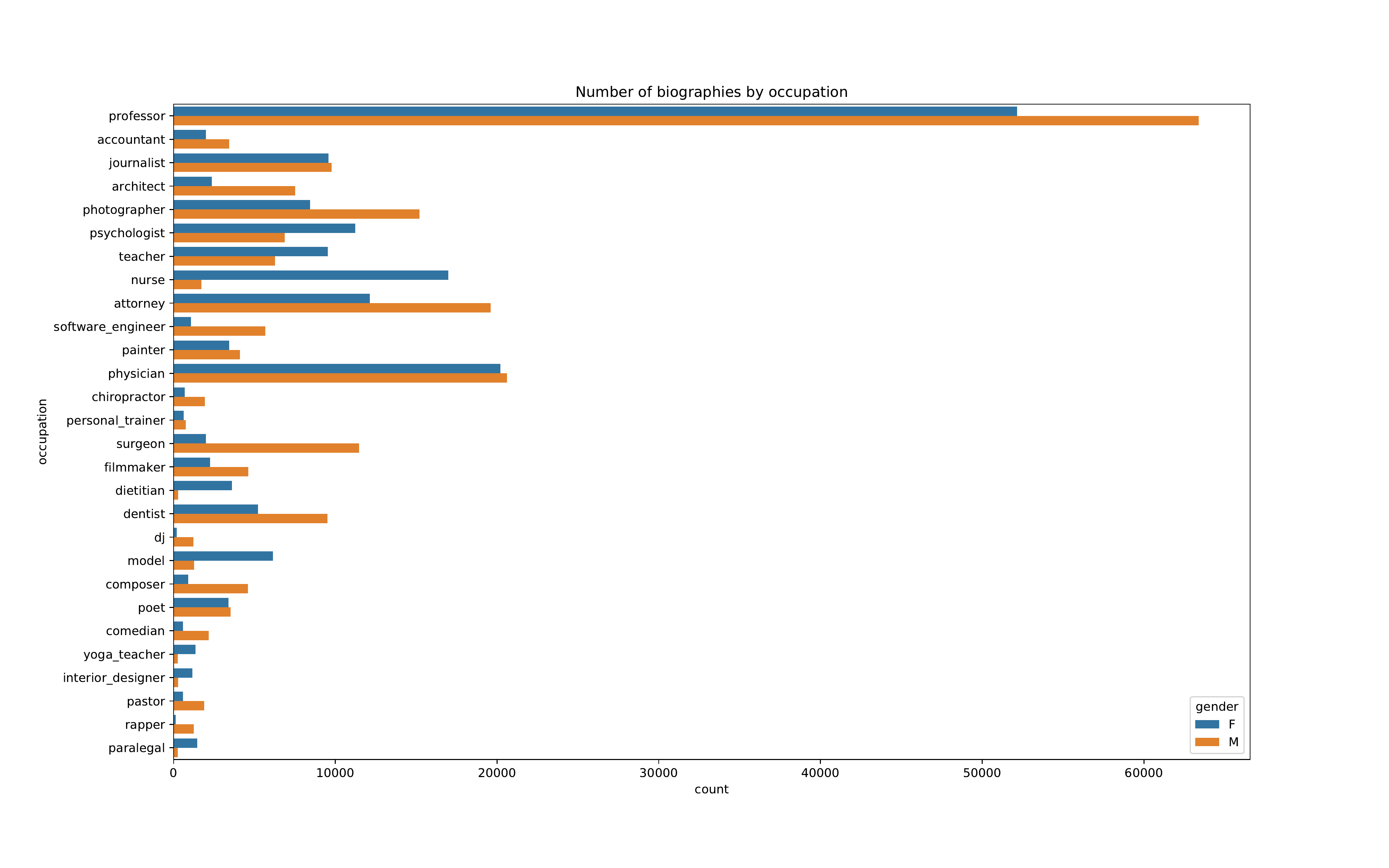}
    \caption{Number of biographies for each occupation by gender on the total \textit{Bios} dataset \cite{de2019bias}.}
    \label{fig:distribution}
\end{figure}

We assess our methodology using the \textit{Bios} \cite{de2019bias} dataset, which  contains about 400K biographies (textual data). For each biography, \textit{Bios} specifies the gender (M or F) of its author as well as its occupation (among 28 occupations, categorical data). As shown in figure \ref{fig:distribution}, this dataset contains heterogeneously represented occupation. Although the representation of some occupations is relatively well-balanced between males and females (\textit{e.g.} professor, journalist, ...), other occupations are particularly unbalanced between males and females (\textit{e.g.} nurse, software engineer, ...).
This dataset is particularly interesting for the fair learning community, as it makes it possible to evaluate how different machine learning strategies can try to predict the true occupations of potential job candidates as accurately as possible, based on their biography, while being as fair as possible when distinguishing males and females.
Note that to build this dataset, its authors used Common Crawl and identified online biographies written in English. Then, they filtered the biographies starting with a name-like pattern followed by the string “is a(n) (xxx) title,” where title is an occupation out of the BLS Standard Occupation Classification system.
Having identified the twenty-eight most frequent occupations, they processed WET files from sixteen distinct crawls from 2014 to 2018, extracting online biographies corresponding to those occupations only. This resulted in about 400K biographies with labelled corresponding occupations.

\subsection{Neural-network model and baseline training strategy}\label{ssec:baselinetraining}

Our task is to predict the occupation using only the textual data of the biography. We do it by using a RoBERTa model \cite{liu2019roberta}, which is based on transformers architecture and is  pretrained with the Masked language modelling (MLM) objective. %
We specifically used a RoBERTa base model pretrained by Hugging Face. All information related to how it was trained can be found in  \cite{DBLP:journals/corr/abs-1907-11692}. It can be remarked, that a very large training dataset was used to pretrain the model, as it was composed of five datasets: \emph{BookCorpus} \cite{moviebook}, a dataset containing 11,038 unpublished books; \emph{English Wikipedia} (excluding lists, tables and headers); \emph{CC-News} \cite{10.1145/3340531.3412762} which contains 63 millions English news articles crawled between September 2016 and February 2019; \emph{OpenWebText} \cite{radford2019language}  an open-source recreation of the WebText dataset  used to train GPT-2; \emph{Stories} \cite{trinh2018simple} a dataset containing a subset of CommonCrawl data filtered to match the story-like style of Winograd schemas. Pre-training was performed on these data, by randomly masking 15\% of the words in each of the input sentences and then trying to predict the masked words
After pre-training RoBERTa parameters on this huge dataset, we then trained it on the 400.000 biographies of the  \emph{Bios} dataset. Training was performed with PyTorch on 2 GPUs (Nvidia Quadro RTX6000 24GB RAM) for 5 epochs with a batch size of 32 observations and a sequence length of 512 words. The optimizer was Adam with a learning rate of 1e-5, $\beta_1 = 0.9$, $\beta_2 = 0.98$, and $\epsilon = 1e6$. Computational times required about 36 hours for each run. We want to emphasize that 5 runs of the training procedure were performed to evaluate the stability of the accuracy and the algorithmic biases. For each of these runs, we the split dataset in 70\% for training, 10\% for validation and 20\% for testing. We will denote as baseline models, the neural-networks trained using this procedure.

\subsection{Evaluating the impact of a gender-neutral dataset}\label{subs:genderneutral}

In order to evaluate the impact of a classic gender unbiasing strategy, we reproduced the baseline training protocol of Section \ref{ssec:baselinetraining} on two \emph{apparently} unbiased versions of the \emph{Bios} dataset.
This classic method for debiasing consists of removing explicit gender indicators ({\em i.e.}{\em 'he', 'she', 'her', 'his', 'him', 'hers', 'himself', 'herself', 'mr', 'mrs', 'ms', 'miss'} and first names).
For a BERT model type, however, we could not just remove words because the model is sensitive to sentence structure, not just lexical information.
We therefore adjusted the method by replacing all the first names by  a neutral first name (\textit{Camille}) and by choosing only one gender for all datasets (e.g., for all individuals of gender g, we did nothing; for the others, we replaced explicit gender indicators with those of g).
We then created two datasets with only female or male gender indicators, and the only first name \textit{Camille}.

Note that by using a fully trained model on our dataset, setting all gender indicators to either feminine or masculine should naively not change anything, since the model would only "know" one gender (which  would therefore be neutral). We however used a pre-trained model on gendered datasets. It is therefore important to verify that fine-tuning this model with a male gendered dataset is equivalent to training it on a female gendered dataset.
To assess this, we carried out several student tests. One between the accuracy of the trained model on the female gendered dataset and the accuracy of the male gendered one. One on the TPR gender gap for each of the professions between the two models. 
None of these tests had a statistically significant difference. We will then only present in Section~\ref{sec:results} the results obtained on the model trained on the female gendered data set.

\subsection{Training procedure for the regularized model}

We now follow the procedure summarized in Algorithm~\ref{alg:BTP_NN_W2_mutli_class} to train bias mitigated multi-class neural-network classifiers on the textual data of the \emph{Bios} dataset.
We first consider the 5  baseline models of Section~\ref{ssec:baselinetraining}, which were trained on the original  \emph{Bios} dataset (and not one of the unbiased datasets of Section~\ref{subs:genderneutral}).
As shown in Fig.~\ref{fig:confmat}-(left), where the diagonal of the confusion matrices differences between males and females represents the TPR gap of all output classes, two classes have TPR gaps above $0.1$ or under $-0.1$: \emph{Surgeon} (in favor of males) and \emph{Model} (in favor of females). We then chose to regularize the predictions for these two occupations.  Note that other occupations could have been considered (see Fig.~\ref{fig:confmat_appendix}) but they contained not enough statistical information to be properly treated. For instance, although the whole training set contains about 400.000 observations, it contains less than 100 female \emph{dj} and less than 100 male \emph{paralegal}.
After having selected these two occupations, we trained 5 regularized models by minimizing Eq.~\eqref{eq:minimizedEnergy_extension}.
We chose a single $\lambda$ parameter for the regularization (the same for both classes, but we could have taken one per class), by using cross validation, with the goal to effectively reduce the TPR gaps on regularized classes without harming the accuracy too much. The best performance/debiasing compromise we found was $\lambda = 0.0001$. An amount of $m=16$ additional observations was used at each mini-batch step to compute each of the discrete cumulative histograms $H_{k,s}$ of the regularization terms pseudo-derivatives Eq.~\eqref{eq:discreteDerivative_extension}. The rest of the training procedure was the same as in Section~\ref{ssec:baselinetraining}. Computational times required about 70 hours for each run.


\section{Results}\label{sec:results}

\begin{figure}
    \centering
    \includegraphics[width=1\linewidth]{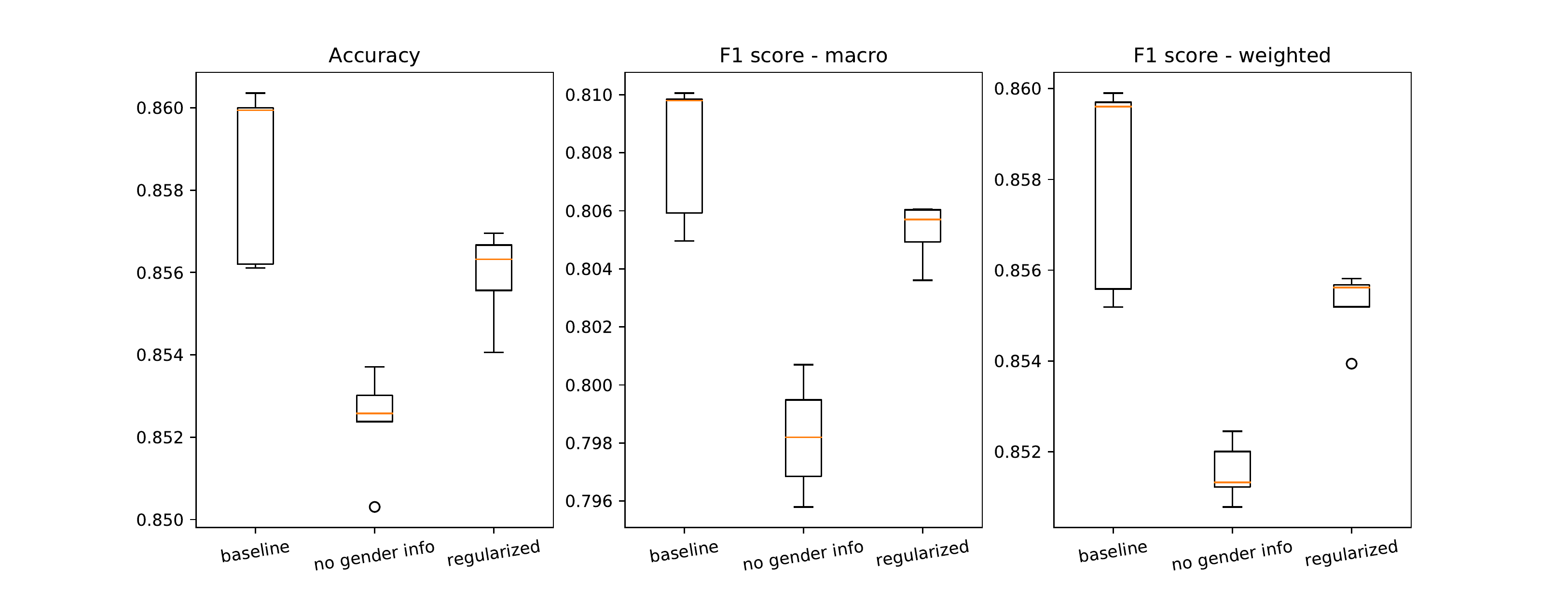}
    \caption{\textbf{(Left to Right)} Box plots of the Accuracy, unweighted F1 scores and weighted F1 scores for the baseline models with raw biographies, the baseline models with unbiased biographies, and the  regularized models with raw biographies.}
    \label{fig:acc_F1scores}
\end{figure}

In commercial applications, fair prediction algorithms will be obviously much more keen to be exploited if they remain accurate. We then made sure that our regularization technique did not had a strongly negative impact on the prediction accuracy. We then quantified different accuracy metrics: First the average accuracy and then two variants of the F1 score, as it is very appropriate for a multiclass classification problem like ours. These two variants are the so-called “macro” F1-score, where we calculate the metric for each class, then we average it without taking into account the number of individuals per class; and the “weighted” F1-score where the means are weighted using the classes representativeness.
We can draw similar conclusions for these three metrics, as shown in Fig.~\ref{fig:acc_F1scores}: our regularization method is certainly a little below the baseline in terms of accuracy, but it is more stable. In addition, it is clearly more accurate than the gender neutralizing technique of Section~\ref{subs:genderneutral}.

\begin{figure}
    \centering
\includegraphics[width=1\linewidth]{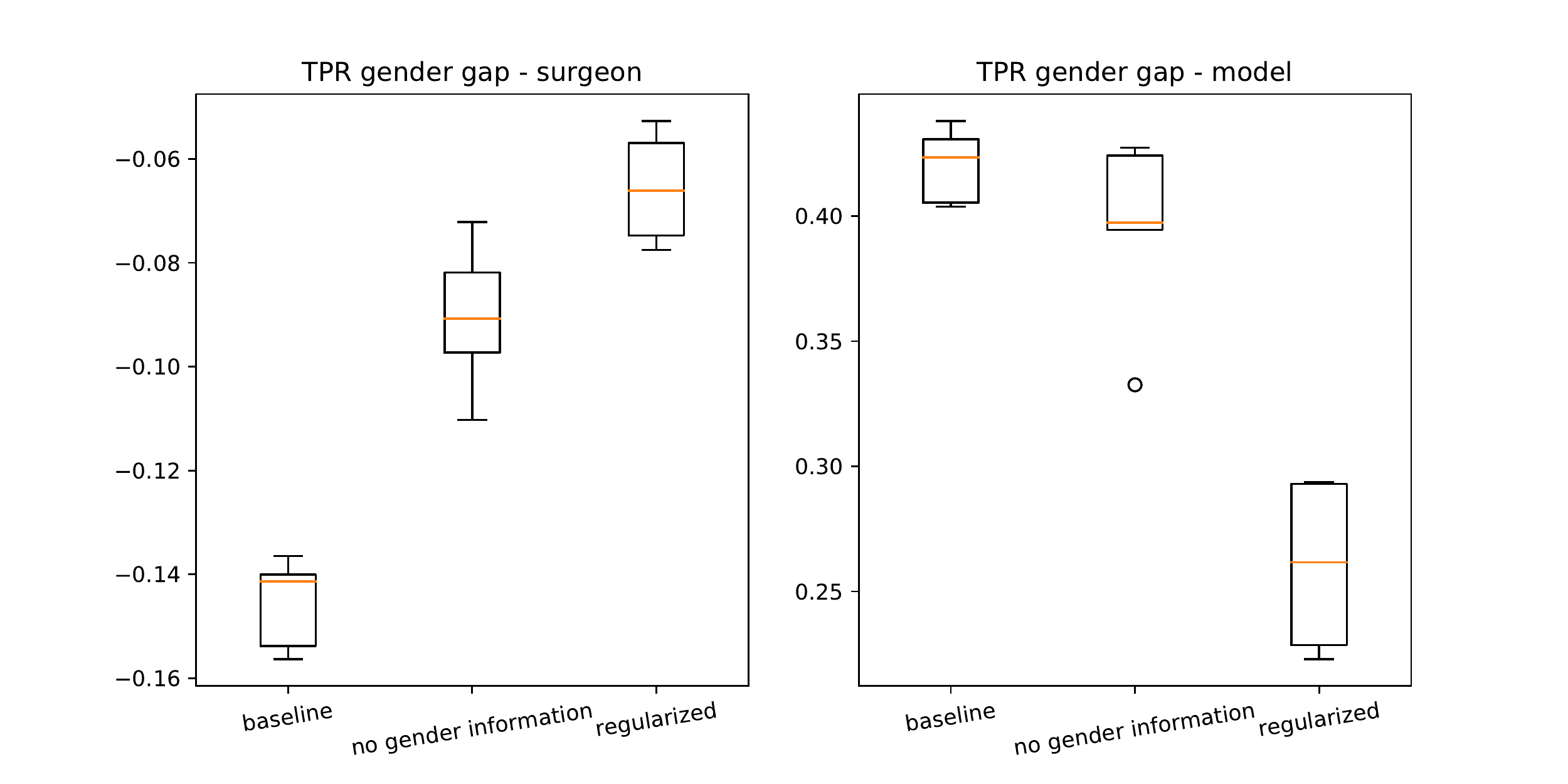}
    \caption{Box plots of the True Positive Rate (TPR) gender gaps for the output classes \textit{Surgeon} and \textit{Model} obtained using the baseline models with raw biographies, the baseline models with unbiased biographies, and the  regularized models with raw biographies. Note that there is a bias in favour of females or males if a TPR gender gap is positive or negative, respectively.}
    \label{fig:TPRGGap}
\end{figure}

We then specifically observed the impact of our regularization strategy in terms of TPR gap on the two regularized classes \emph{Surgeon} and \emph{Model}. Boxplots of the TPR gap for these output classes are shown in  Fig.~\ref{fig:TPRGGap}. They  confirm that the algorithmic bias has been reduced for these two classes. For the class \emph{Surgeon}, removing gender indicators had a strong effect, but the regularization strategy further reduced the biases.  For the class \emph{Model}, removing gender indicators had little effect, and the regularization strategy reduced the biases by almost a factor two.

\begin{figure}
    \centering
\includegraphics[width=1.0\linewidth]{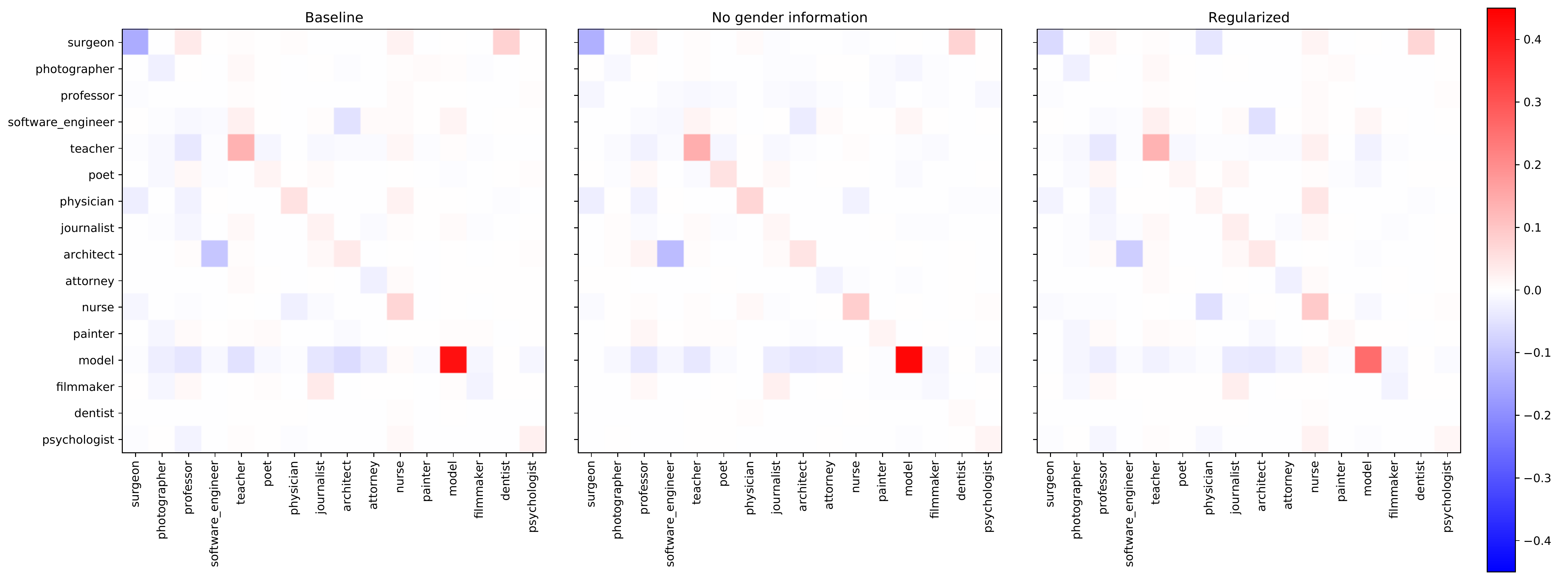}
    \caption{Average difference between the confusion matrices of the predicted outputs $\hat{y}$ versus the true outputs $y$  obtained for females and males. Note that the diagonal values of these matrices correspond to average TPR gaps. The confusion matrices were also normalized over the true (rows) conditions. The redder a value the stronger  the bias in favour of females, and the bluer a value the stronger the bias in favour of males.}
    \label{fig:confmat}
\end{figure}

\begin{figure}
    \centering
\includegraphics[width=1\linewidth]{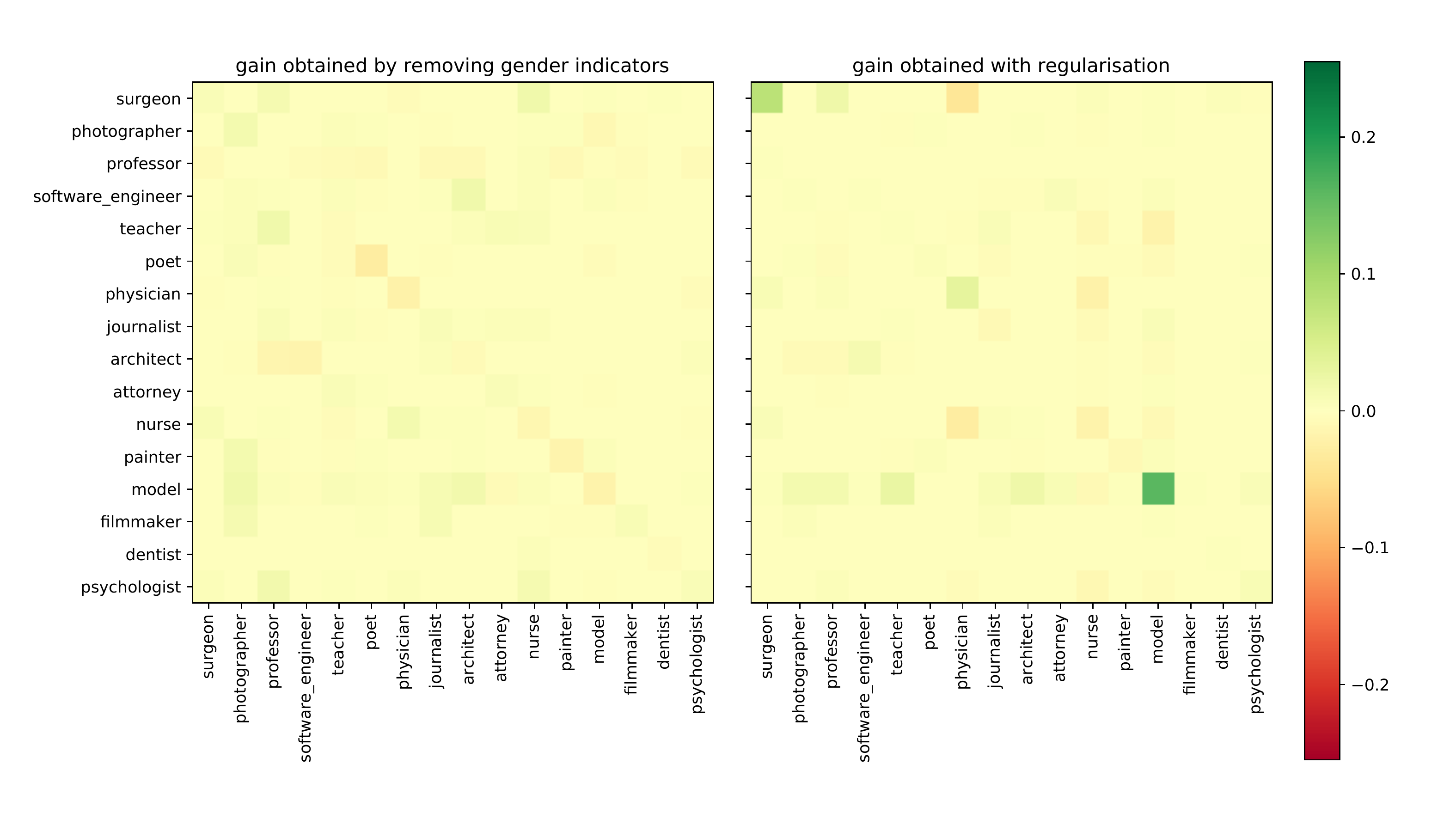}
    \caption{\textbf{(Left)} Difference between the baseline matrix of Fig.~\ref{fig:confmat}, and the one obtained with an unbiased training set. \textbf{(Right)} Difference between the baseline matrix of Fig.~\ref{fig:confmat}, and the one obtained using regularized optimization. The greener the values, the more the technique has reduced the bias between men and women. The redder the values, the more the bias has been amplified.}
    \label{fig:confmat_gains}
\end{figure}

We finally wanted to make sure that reducing the unacceptable biases on these two classes would not be at the expense of newly generated biases. We then measured the difference between the average (on the 5 models) confusion matrix for females and males only.
In Fig.~\ref{fig:confmat}, we see the evolution of our biases according to the selected method.
Note first that the diagonal of these matrix differences corresponds to the TPR gaps.
Remark too that we only represent the results obtained on the 16 most frequent occupations for visibility concerns, but  are the complete matrices are show in the appendix.
On our two regularized classes, we are getting closer to white (i.e. non-bias), and for the other classes, we also observe a decrease in bias in general, and no outlier point.
For a finer analysis and more clarity, we represent in Fig.~\ref{fig:confmat_gains}  the difference between the absolute values of the baseline matrix of Fig.~\ref{fig:confmat} and each of the compared matrices (\emph{i.e.} with neutralized genders and regularization).
This clearly represents us the “gains” of these two bias reduction methods to compare them. Figure \ref{fig:confmat_gains} confirms our intuition given in Fig.~\ref{fig:confmat}: in the case where the gender indicators are removed, the gain is rather slight and depends on the class. In the case of our regularization, the two regularized classes obtain a very clear positive gain, and there is no marked negative gain on the rest of the matrix.

\section{Discussion}

In this paper, we have defined a strategy to address the critical need for certifying that commercial prediction models present moderate discrimination biases.  We specifically defined a new algorithm to mitigate undesirable algorithmic biases in multi-class neural-network classifiers, and applied it to NLP application that is ranked as \emph{High risk} by E.U. regulations. Our method was shown to successfully temper algorithmic biases in this application, and outperformed a classic strategy both in terms of prediction accuracy and mitigated bias. In addition, computational times were only reasonably increased compared with a baseline training method.
The state of the art of \textit{in-processing} unbiasing methods mainly focuses on binary models, and our approach addresses the multiclass problem. The possibility of choosing which classes to regularize and of applying a different $\lambda$ for each class gives a wide range of application of the method.

We finally want to emphasize that although our method was applied to NLP data, it can be easily applied to any multi-class neural-network classifier. We also believe that it could be simply adapted to other fairness metrics. Our regularization method is implemented to work as a loss in PyTorch and is compatible with PyTorch-GPU. It is freely available on GitHub\footnote{\url{https://github.com/lrisser/W2reg} -- The binary classification regularizer for images and tabular data is currently distributed. The multi-class extension for NLP data, images, and tabular data will also be distributed, subject to paper acceptance}.

\section*{Acknowledgments}
    This research was funded by the AI (Artificial Intelligence) Interdisciplinary Institute ANITI (Artificial and Natural InTelligence Institute.), which is funded by the French ‘Investing for the Future– PIA3’ program under the Grant agreement ANR-19-PI3A-0004.

\appendix

\section{Results obtained on all output classes}

\begin{figure}
    \centering
\includegraphics[width=1.\linewidth]{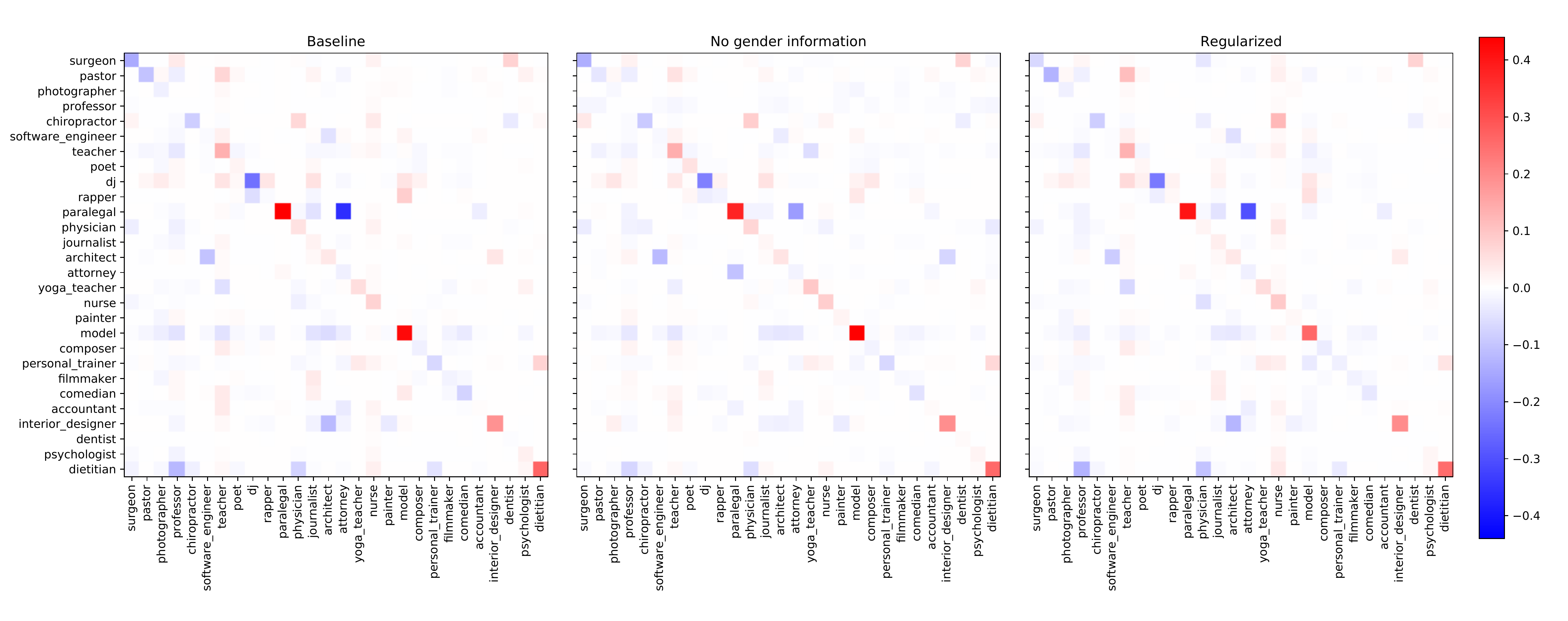}
    \caption{Extension of Fig.~\ref{fig:confmat} to all output classes of the Bios dataset and not only the most frequent ones, which were selected in Fig.~\ref{fig:confmat} for readability purposes.}
    \label{fig:confmat_appendix}
\end{figure}

\begin{figure}
    \centering
\includegraphics[width=1\linewidth]{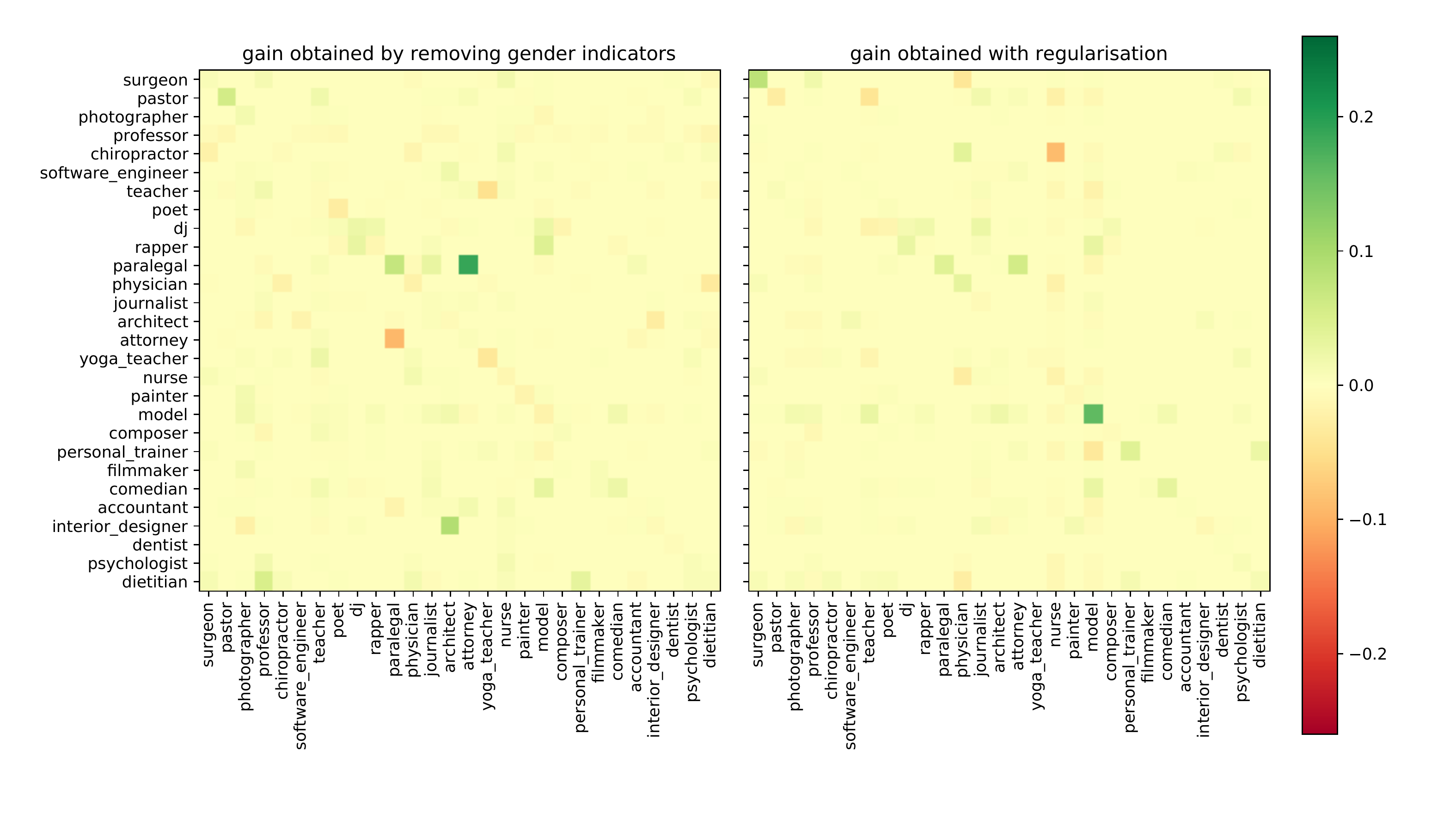}
    \caption{Extension of Fig.~\ref{fig:confmat_gains} to all output classes of the Bios dataset and not only the most frequent ones, which were selected in Fig.~\ref{fig:confmat_gains} for readability purposes.}
    \label{fig:confmat_gainsall}
\end{figure}

The results shown in Figs.~\ref{fig:confmat} and \ref{fig:confmat_gains} selected the most largely represented output classes for readability purposes. We show in this appendix their extensions, Figs.~\ref{fig:confmat_appendix} and \ref{fig:confmat_gainsall}, to all output classes of the \textit{Bios} dataset \cite{de2019bias}. It can be observed in these figures that other output classes than \textit{Model} and \textit{Surgeon} presented high gender biases, when using the baseline strategy: \textit{Paralegal}, \textit{DJ} and \textit{Dietician}. Although we used these output classes when training the prediction model to make the classification task complex, we voluntarily decided to not regularize them for statistical concerns:
These occupations are indeed first poorly represented in the \textit{Bios} dataset and are additionally strongly unbalanced between males and females. Although the whole training set contains more than 400,000 biographies, there are less than 100 biographies for female \textit{DJ}, male \textit{Dietician} and male \textit{Paralegal}. This makes their treatment with a statistically-sound strategy unreliable. When applied to statistically poorly represented observations, a constrained neural-network won't indeed learn to use generalizable features in the input biographies, but will instead overfit the specificities of each observation which is strongly highlighted by the constraint. We can however see that the tested bias mitigation strategies on the classes \textit{Model} and \textit{Surgeon} did not amplify the biases on the \textit{Paralegal}, \textit{DJ} and \textit{Dietician} classes.

From a certification perspective in the E.U., the \emph{AI act} will ask to clearly mention to end-users the cases for which the predictions may be unreliable or potentially biased. In this context, our strategy makes it possible to certify that mutli-class neural-network classifiers make unbiased decisions on output classes that would be biased using standard training, if the training data offer a sufficient representativity and variability of the characteristics in these classes. In the case where a company would desire to certify that poorly represented classes in the training set are free of biases, the certification procedure will naturally require acquiring more observations.

\end{document}